\documentclass[10pt,letterpaper]{article}

\usepackage{cogsci}

\cogscifinalcopy %%% Uncomment this line for the final submission

\usepackage{cmap}
\usepackage[T1]{fontenc}
\usepackage[american]{babel}
\usepackage{csquotes}
\usepackage{newtxtext,newtxmath}  %%% TeX Gyre Termes

\usepackage[
  backend=biber,
  style=apa,
  natbib=true,
  eprint=true,
  annotation=false,
]{biblatex}
\usepackage{graphicx}
\usepackage{float} %%% Roger Levy added this and changed figure/table placement to [H] for conformity to Word template, though floating tables and figures to top is still generally recommended!

\usepackage[hidelinks]{hyperref}
\usepackage{cleveref}
\usepackage{booktabs}

\title{N-gram-like Language Models Predict Naturalistic Reading Time Best}

\author[1,2]{\mbox{James A. Michaelov (jamic@mit.edu)}}
\author[1]{\mbox{Roger P. Levy (rplevy@mit.edu)}}
\affil[1]{Department of Brain and Cognitive Sciences, Massachusetts Institute of Technology}
\affil[2]{MIT Libraries CREOS, Massachusetts Institute of Technology}

\begin{document}

\maketitle

\begin{abstract}
Recent work has found that contemporary language models such as transformers can become so good at next-word prediction that the probabilities they calculate become worse for predicting naturalistic reading time. In this paper, we propose that this can be explained by reading time being shaped by simple $n$-gram statistics rather than the more complex statistics learned by state-of-the-art transformer language models. We demonstrate that the neural language models whose predictions are most correlated with $n$-gram probability are also those that calculate probabilities that are the most correlated with eye-tracking-based metrics of reading time on naturalistic text.

\end{abstract}

\section{Introduction}

Two decades ago, researchers discovered that a word's bigram probability---that is, the probability of the word occurring given a specific previous word---is significantly correlated with the time taken to read the word, with higher-probability words being read faster than lower-probability words \citep{mcdonald_2003_EyeMovementsReveal,mcdonald_2003_LowlevelPredictiveInference}. In the years since, evidence has continued to mount that bigram surprisal (negative log-probability; see \citealp{hale_2001_ProbabilisticEarleyParser}) is a reliable predictor of reading time, as is surprisal from higher-order $n$-grams where probability is calculated based on the $n-1$ previous words rather than the just the previous word \citep{levy_2008_ExpectationbasedSyntacticComprehension,smith_2008_OptimalProcessingTimes,smith_2011_ClozeNoCigar,smith_2013_EffectWordPredictability,boston_2008_ParsingCostsPredictors,demberg_2008_DataEyetrackingCorpora,mitchell_2010_SyntacticSemanticFactors,frank_2011_InsensitivityHumanSentenceProcessing,fossum_2012_SequentialVsHierarchical,brothers_2021_WordPredictabilityEffects,oh_2022_ComparisonStructuralParsers}. As natural language technology has advanced, researchers have increasingly turned to more complex language models to calculate statistical probability, including recurrent neural networks \citep{frank_2011_InsensitivityHumanSentenceProcessing,fossum_2012_SequentialVsHierarchical,monsalve_2012_LexicalSurprisalGeneral,goodkind_2018_PredictivePowerWorda,aurnhammer_2019_ComparingGatedSimple,aurnhammer_2019_EvaluatingInformationtheoreticMeasures,eisape_2020_ClozeDistillationImproving,hao_2020_ProbabilisticPredictionsPeople,wilcox_2020_PredictivePowerNeural,merkx_2021_HumanSentenceProcessing,brothers_2021_WordPredictabilityEffects,kuribayashi_2021_LowerPerplexityNot,oh_2022_ComparisonStructuralParsers} and more recently, transformers \citep{merkx_2021_HumanSentenceProcessing,hao_2020_ProbabilisticPredictionsPeople,wilcox_2020_PredictivePowerNeural,wilcox_2023_TestingPredictionsSurprisal,wilcox_2023_LanguageModelQuality,eisape_2020_ClozeDistillationImproving,kuribayashi_2021_LowerPerplexityNot,oh_2022_ComparisonStructuralParsers,oh_2023_WhyDoesSurprisal,oh_2023_TransformerBasedLanguageModel,oh_2024_FrequencyExplainsInverse,oh_2025_InverseScalingEffect,oh_2025_DissociableFrequencyEffects,shain_2024_WordFrequencyPredictability,shain_2024_LargescaleEvidenceLogarithmic,opedal_2024_RoleContextReading,meister_2021_RevisitingUniformInformation}.

An initial trend reported in this line of work was that more powerful models---those with more parameters, those with more advanced or scalable architectures, those trained on larger corpora, and those with better next-word prediction capabilities---produce predictions that show the best fit to human reading times for naturalistic text \citep{goodkind_2018_PredictivePowerWorda,wilcox_2020_PredictivePowerNeural,wilcox_2023_LanguageModelQuality}. For example, \citet{goodkind_2018_PredictivePowerWorda} observed a linear relationship between a language model's perplexity (a measure of next-word prediction capability; \citealp{jelinek_1977_PerplexityMeasureDifficulty}) and the fit between its predictions and reading time data.

However, this pattern has not remained robust \citep{eisape_2020_ClozeDistillationImproving,hao_2020_ProbabilisticPredictionsPeople,merkx_2021_HumanSentenceProcessing}, and the evidence has begun to accumulate for the existence of a limit to the extent to which this is the case: contemporary transformer language models can become so good at next-word prediction that the surprisals they assign to words become less correlated with naturalistic reading time \citep{kuribayashi_2021_LowerPerplexityNot,oh_2022_ComparisonStructuralParsers,oh_2023_WhyDoesSurprisal,oh_2023_TransformerBasedLanguageModel,oh_2024_FrequencyExplainsInverse,shain_2024_LargescaleEvidenceLogarithmic}. This \textit{inverse scaling} effect does not appear to be determined by any specific factor that makes a given model better at next-word prediction; it applies to cases where larger models (i.e., those with a greater number of parameters) are better at next-word prediction than smaller models \citep{oh_2022_ComparisonStructuralParsers,oh_2024_FrequencyExplainsInverse}, cases where models trained on larger datasets are better at next-word prediction than models trained on smaller datasets \citep{oh_2023_TransformerBasedLanguageModel}, and possibly cases where some architectures tend to perform better at next-word prediction than others \citep{eisape_2020_ClozeDistillationImproving}.

Other limitations of relying solely on neural language models to calculate surprisal have also emerged. Unigram surprisal (i.e., negative log-frequency) is often a significant predictor of reading time above and beyond language model surprisal \citep{shain_2024_WordFrequencyPredictability}, despite the latter being statistical models of language probability. The same may also be true for bigram and trigram surprisal \citep{goodkind_2021_LocalWordStatistics}. In fact, the extent to which adding unigram and bigram surprisal as predictors improves the fit of a regression predicting reading time increases as models become better next-word predictors, which has been interpreted as `compensation' for the increasingly worse fit of neural language model surprisal \citep{oh_2025_DissociableFrequencyEffects}. In addition, much of the effect of surprisal on reading time appears to be explainable by frequency alone \citep{opedal_2024_RoleContextReading}. Thus, the evidence suggests that contemporary language model surprisal does not fully capture the effects of statistical probability on reading time, and as models become larger and more powerful, their surprisals do so less. 

\subsection{Where does the inverse scaling come from?}
\citet{oh_2023_TransformerBasedLanguageModel} find that using surprisal from language models trained on up to 2 billion words leads to predictions that better fit naturalistic reading time data, but surprisal from models trained beyond this point shows an increasingly worse fit to the data. This pattern is also observable when comparing models according to their perplexity---models appear to become `too good' at next-word prediction for their surprisals to fit reading time well \citep{oh_2023_TransformerBasedLanguageModel}. One specific observed phenomenon is that more powerful models assign relatively lower surprisals to low-frequency words (especially nouns and adjectives; \citealp{oh_2023_WhyDoesSurprisal}) than do less powerful models, and this in turn can lead to a systematic under-estimation of these words' reading times \citep{oh_2024_FrequencyExplainsInverse}. This may also partially explain why frequency has been found to explain variance in reading time above and beyond surprisal \citep{shain_2024_WordFrequencyPredictability,oh_2025_DissociableFrequencyEffects}. However, exactly what underlies the decreasing fit between language model predictions and human language processing as indexed by reading time is still not fully understood.

A general explanation for these patterns is that during the earlier phases of training, language models make increasingly human-like predictions as they initially begin to learn language statistics, but as they continue to train, their predictions increasingly reflect properties of the data and their own inductive biases. One version of this is the aforementioned possibility that models simply become `too good' at next-word prediction overall due to being trained on orders of magnitude more data than humans are exposed to, thereby being able to predict low-frequency words when humans may not be \citep{oh_2024_FrequencyExplainsInverse}. But data type could also matter. For example, during early language acquisition, our input is primarily made up of speech, and often child-directed speech. by contrast, language models are trained on `high-quality' text data, selected based on specific criteria that are not aligned with this \citep[see, e.g.,][]{wilcox_2025_BiggerNotAlways}. It is also possible that even if the models were trained on exactly the same data that the average human were exposed to, the predictions of humans and language models would begin to diverge as the models better learned to reflect the empirical distribution---previous work suggests that human predictions may not always exactly reflect empirical probabilities, potentially due to our own inductive biases or the augmentation of predictions with other kinds of knowledge \citep[see, e.g.,][]{smith_2011_ClozeNoCigar,brothers_2021_WordPredictabilityEffects,michaelov_2022_ClozeFarN400,michaelov_2024_MathematicalRelationshipContextual}. Finally, different model architectures are more or less cognitively plausible in a range of ways, which could in principle impact the extent to which their predictions match our own \citep{merkx_2021_HumanSentenceProcessing,michaelov_2021_DifferentKindsCognitive,ryu_2021_AccountingAgreementPhenomena}, though no impact of this on the inverse scaling effect has been observed \citep{michaelov_2024_RevengeFallenRecurrent}.

\begin{figure*}[ht]
    \centering
    \includegraphics[width=0.9\linewidth]{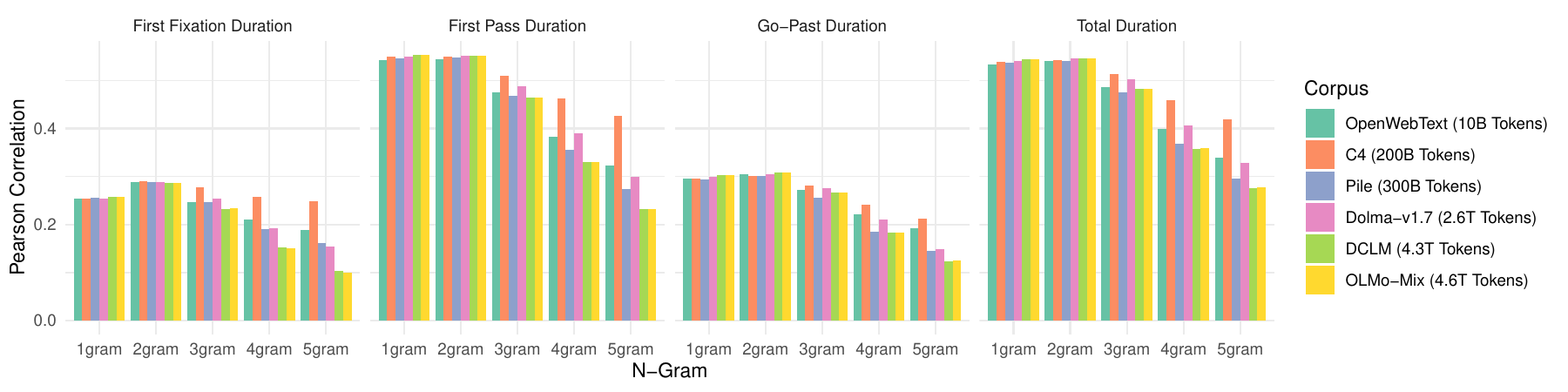}
    \caption{The correlation between $n$-gram surprisal and reading time in the Provo Corpus.}
    \label{fig:provo_ngram}
\end{figure*}

\subsection{The $n$-gram hypothesis} 
While the aforementioned possibilities are plausible and likely worthwhile to explore further, in this paper we propose a different explanation, which we empirically evaluate on eye-tracking data: \textbf{naturalistic reading time is shaped by lower-order $n$-gram statistics, and thus, language models that make the most $n$-gram-like predictions show the best fit to the data}. 

This would account for the aforementioned inverse scaling patterns observed in previous work, aligns with older work using $n$-gram surprisal to predict reading time, and is supported by several additional findings. First, researchers have often found that operations that reduce the role of context---including restricting the context window \citep{kuribayashi_2022_ContextLimitationsMake}, reducing memory capacity \citep{timkey_2023_LanguageModelLimited}, or adding a recency bias \citep{goodkind_2021_LocalWordStatistics,vaidya_2023_HumansLanguageModelsa,devarda_2024_LocallyBiasedTransformers,clark_2025_LinearRecencyBias}---improve language model predictions' fit to reading time. In addition, language models of all architectures tend to show a consistent training trajectory where their predictions closely match unigram probability, followed by bigram probability, trigram probability, and so on \citep{karpathy_2016_VisualizingUnderstandingRecurrent,chang_2024_CharacterizingLearningCurves,michaelov_2025_LanguageModelBehavioral}. When \citet{oh_2023_TransformerBasedLanguageModel} looked at the Pythia suite of language models \citep{biderman_2023_PythiaSuiteAnalyzing}, they found that fit to reading time increases until step 1000 of training (2 billion training tokens), at which point the fit begins to decrease. More recently, \citet{michaelov_2025_LanguageModelBehavioral} found that step 1000 is around the point during training at which the Pythia models' predictions' correlations to bigram and trigram probabilities peak; after which their correlation drops, as does the correlation with unigram probability. This is exactly what one would expect if reading time aligns with these $n$-gram statistics.

Why $n$-grams? Our proposal is derived from considering the time course of the neurocognitive mechanisms at play during naturalistic reading. Current research on the neuroscience of language comprehension suggests that the N400 component of the event-related brain potential (ERP), which reliably occurs 300-500ms after a stimulus is first encountered, indexes the activation of a word's representations in the brain \citep{kutas_2011_ThirtyYearsCounting,brouwer_2012_GettingRealSemantica,kuperberg_2020_TaleTwoPositivities,federmeier_2021_ConnectingConsideringElectrophysiology,aurnhammer_2021_RetrievalN400Integration}, and is followed by a number of positive ERP components that have been associated with integration, reanalysis, or anomaly detection and correction \citep{brouwer_2012_GettingRealSemantica,delong_2020_ComprehendingSurprisingSentences,kuperberg_2020_TaleTwoPositivities,aurnhammer_2021_RetrievalN400Integration}. But in naturalistic reading, the average time spent on a word during the first pass (i.e., first pass/gaze duration) is generally under 250ms \citep{kennedy_2013_FrequencyPredictabilityEffects,cop_2017_PresentingGECOEyetracking,berzak_2025_OneStop360ParticipantEnglish}. Taking the lower bound of estimates about how long is needed to program the motor actions of a saccade (110-150ms; \citealp{becker_1979_AnalysisSaccadicSystem,reichle_2012_EyeMovementsReading}) and the upper bound of estimates of how long a saccade takes to complete on average during natural reading (20-40ms; \citealp{rayner_1981_EyeMovementsIdentifying,reichle_2012_EyeMovementsReading}), if the first pass duration of a word is 250ms, this would put the initiation of the programming of the saccade from word $w_i$ to $w_{i+1}$ at a maximum of 450ms after word $w_{i-1}$ was first fixated; which is during the N400, and well before any of the later processes have even begun. Thus, it is hard to see how the first pass duration of the average word $w_{i}$ could be shaped by a fully-integrated representation of $w_1...w_{i-1}$. To the extent that reading time is shaped by predictions about upcoming words, these predictions must, at least some of the time, be based on shallower, under-specified, or otherwise incomplete representations of their context.

One strong potential candidate for this is $n$-gram prediction. Prediction based on $n$-grams relies only on having identified the previous words in the sequence. Under a hierarchical predictive coding model of language comprehension \citep[e.g.,][]{lewis_2015_PredictiveCodingFramework,bornkessel-schlesewsky_2019_NeurobiologicallyPlausibleModel,kuperberg_2020_TaleTwoPositivities,noureddine_2024_PredictiveCodingModel}, for example, this could be accomplished relatively early by the lower levels of the predictive system that are associated with word form at the same time as the N400 and later ERP components begin to unfold at higher levels. And the fact that recurrent neural networks, transformers, and other modern language models, all of which are explicitly optimized for next-word prediction also all learn $n$-gram relations \citep{karpathy_2016_VisualizingUnderstandingRecurrent,sun_2022_ImplicitNgramsInduced,choshen_2022_GrammarLearningTrajectoriesNeural,bietti_2023_BirthTransformerMemory,voita_2024_NeuronsLargeLanguage,akyurek_2024_InContextLanguageLearning,chang_2024_CharacterizingLearningCurves,chang_2025_BigramSubnetworksMapping,michaelov_2025_LanguageModelBehavioral} may be taken to suggest that it is rational for any predictive system to learn such patterns. It is therefore plausible that $n$-gram and potentially other shallow statistics could account for many of the predictability effects observed in naturalistic reading.

It is worth noting that the $n$-gram hypothesis is not intended to be an account of \textit{all} contextual or probabilistic reading time effects. Longer-range dependencies are known to shape processing \citep{staub_2006_SyntacticPredictionLanguage}; and indeed, our account does not preclude the possible impact of rich contextual information from $w_{<i-1}$. But since natural texts---which are the context in which inverse scaling effects have been observed---are generally written to be coherent rather than explicitly oriented around known loci of processing difficulty (as in the case of psycholinguistic stimuli), it is likely that this does not have as much of a measurable impact above and beyond surface-level statistics operationalized by $n$-grams. It also worth noting that our account is primarily focused on first pass duration, but the inverse scaling effect has also been observed for go-past duration \citep{oh_2022_ComparisonStructuralParsers,oh_2023_WhyDoesSurprisal,oh_2023_TransformerBasedLanguageModel,oh_2024_FrequencyExplainsInverse,oh_2025_InverseScalingEffect,oh_2025_DissociableFrequencyEffects}. But since go-past duration is largely made up of first pass duration, it would be surprising if the former showed a different scaling pattern unless there were some specific reason to believe that surprisal from more powerful language models would be sufficiently better at predicting the time spent on regressions out of words to counteract the worse prediction of first pass duration. We are not aware of any direct evidence for such a prediction; and in fact, what causes regressions is still an open research question \citep{wilcox_2024_InformationtheoreticAnalysisTargeted,rego_2026_WhatDrivesRegressions,clark_2026_ReadersMakeTargeted}.

\begin{figure*}[ht]
    \centering
    \includegraphics[width=0.9\linewidth]{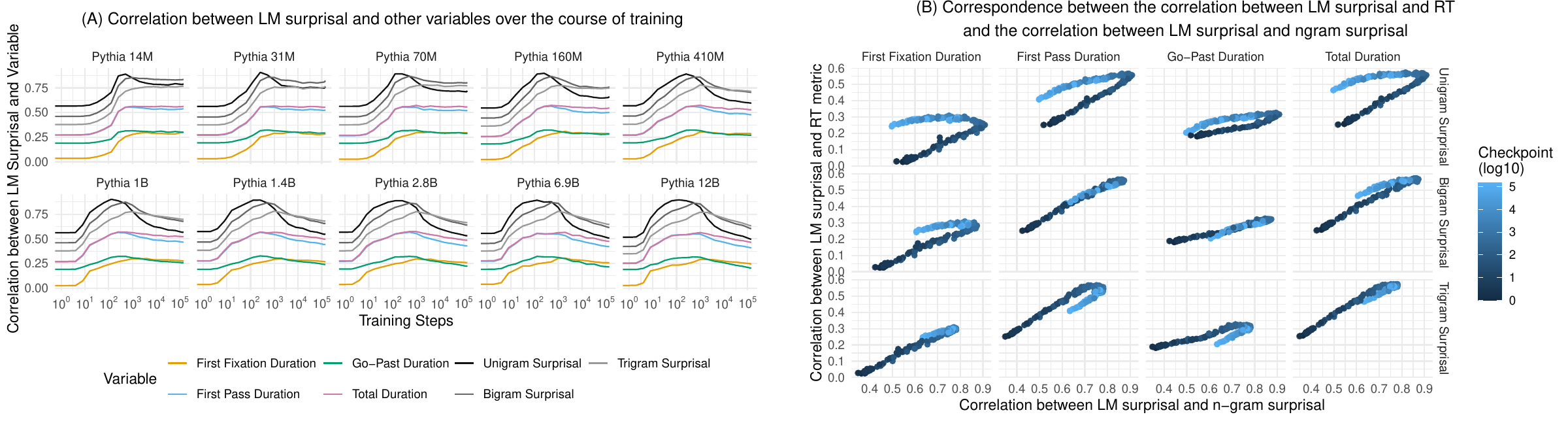}
    \caption{(A) The relationship between language model surprisal over the course of training and both $n$-gram surprisal and reading time in the Provo Corpus. (B) The relationship between the two sets of correlations.}
    \label{fig:corr_provo}
\end{figure*}

\section{Experiment 1}
If the $n$-gram hypothesis holds, we should expect the relationship between $n$-gram surprisal and reading time to show a different pattern to that of neural language models. Crucially, if reading time reflects lower-order $n$-gram relations, then these should not show the same inverse scaling. That is, the correlation between lower-order $n$-gram surprisal and reading time should not get worse no matter how large the corpus---if anything, it should improve. Meanwhile, we might expect to see a decrease in correlation as $n$-gram order increases. This is not the pattern observed in past work \citep{goodkind_2018_PredictivePowerWorda,hao_2020_ProbabilisticPredictionsPeople}, but it is worth noting that the $n$-grams in these studies relied on corpora of roughly 1B tokens, which is before inverse scaling emerges in transformers \citep{oh_2023_TransformerBasedLanguageModel}. We carry out this analysis with contemporary corpora ranging in size from tens of billions to trillions of tokens.

\subsection{Method}
\subsubsection{$n$-grams}
We calculate the surprisal (negative log-probability) of words based on their $n$-gram probability in six corpora: OpenWebText \citep{gokaslan_2019_OpenWebTextCorpus}, C4 \citep{raffel_2020_ExploringLimitsTransfer}, the Pile \citep{gao_2020_Pile800GBDataset}, Dolma \citep{soldaini_2024_DolmaOpenCorpusa}, DCLM \citep{li_2024_DataCompLMSearchNext}, and OLMo-Mix \citep{olmoteam_2025_2OLMo2a}. Following \citet{michaelov_2025_LanguageModelBehavioral}, we calculate $n$-gram probability using the raw counts of word sequences in the corpus as calculated by \textit{infini-gram} \citep{liu_2024_InfinigramScalingUnbounded}, and use the Stupid Backoff \citep{brants_2007_LargeLanguageModels} procedure to convert these counts into probabilities. For OpenWebText $n$-grams, we build an infini-gram index locally; for the other corpora, we rely on the infini-gram API. We provide all data, code and analysis scripts at \url{https://osf.io/msuwk}.

\subsubsection{Reading Time}
We carry out our analysis on the Provo Corpus \citep{luke_2018_ProvoCorpusLarge}, an eye-tracking dataset comprised of data from 470 participants reading 55 passages. We consider four reading time measures: First Fixation Duration, First Pass Duration (duration from when a word is first fixated to when it is first left), Go-Past Duration (duration from when a word is first fixated to the first fixation \textit{past} the word), and Total Duration (the sum of all fixations on the word). We exclude words that were skipped on the first pass, as well as words at the beginning of end of a sentence (see, e.g., \citealp{oh_2023_TransformerBasedLanguageModel,shain_2024_LargescaleEvidenceLogarithmic}). We then calculate the mean of each metric for each word in the corpus (2,449 after the aforementioned exclusions). Of the aforementioned metrics, we are only aware of one previous study investigating scaling effects on the prediction of first fixation duration and total duration \citep{hao_2020_ProbabilisticPredictionsPeople}.

\subsection{Results}
The results of this experiment are presented in \Cref{fig:provo_ngram}. The first clear pattern we observe is that unigram and bigram surprisal tend to be most highly correlated with reading time with little difference between the two (the one exception to this is First Fixation Duration, where bigram surprisal has a substantially higher correlation). Each $n$-gram beyond this (i.e., for $n\geq3$) shows a progressively lower correlation to each reading time metric, and in addition, for these $n$-grams, there is a general tendency for $n$-grams based on larger corpora to show a lower correlation. Finally, we see corpus-specific effects. Specifically, we see that C4 $n$-grams, and to a lesser extent Dolma $n$-grams, are more highly correlated with reading time than the two aforementioned patterns would predict.

\subsection{Discussion}
We find that, across the board, lower-order $n$-grams ($n\in\{1,2\}$) show a stronger correlation to reading time metrics than higher-order $n$-grams ($n\in\{3,4,5\}$), consistent with our predictions. While we do observe some inverse scaling as a function of dataset size, this only occurs for the higher-order $n$-grams, and so may simply be due to these larger datasets allowing for better estimation of these higher-order $n$-grams. For the lower-order $n$-grams, if anything, there is a very small but noticeable degree of positive scaling as a function of corpus size.

Additionally, we observe that there is substantial and apparently idiosyncratic variation between $n$-grams calculated on different corpora, even after accounting for corpus size. One possible explanation for this is text domain or genre. For example, the fact that C4 was constructed from cleaned Common Crawl data, which includes data from the whole internet may mean that it has a more naturalistic distribution than the others, which are designed explicitly to be `high-quality' data that lead to good language model performance---this includes the intentional inclusion of books and scientific articles, only collecting webpages that have been shared by users on websites such as Reddit, and aggressive data filtering procedures.

\begin{figure*}[ht]
    \centering
    \includegraphics[width=0.9\linewidth]{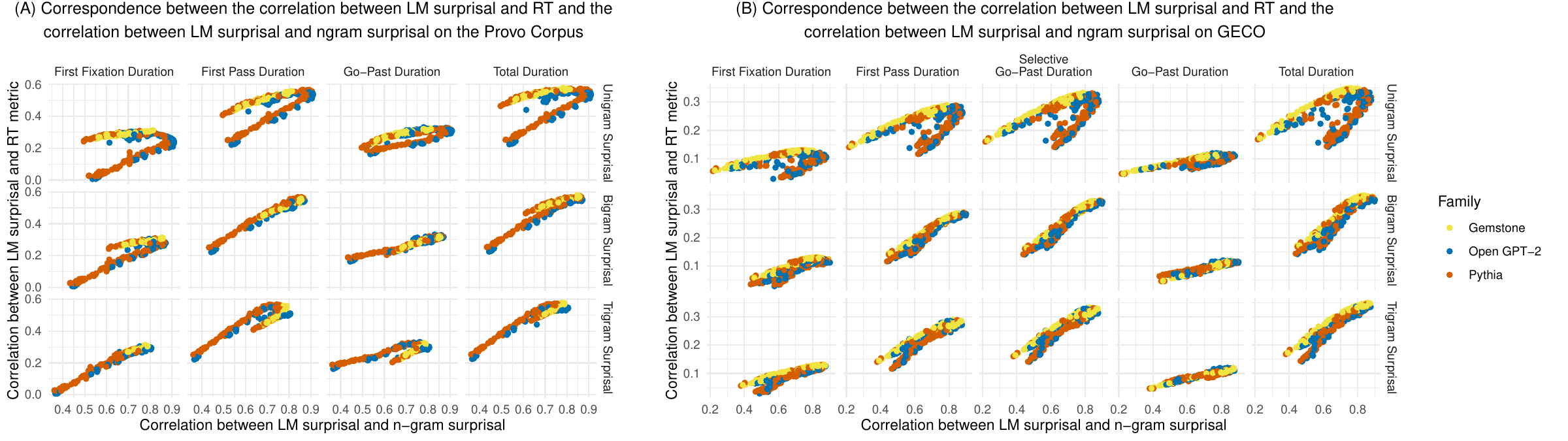}
    \caption{The relationship between the correlation between language model surprisal and $n$-gram surprisal and the correlation between language model surprisal and each measure of reading time in the (A) Provo and (B) GECO datasets.}
    \label{fig:et_both}
\end{figure*}

\section{Experiment 2}
We next turn to our core question of whether the extent to which scaling effects observed in the relationship between language models and reading time can be explained by the extent to which language model surprisal resembles that of lower-order $n$-grams, as would be expected if reading time is sensitive to $n$-gram probability. First, we attempt to replicate the results of \citet{oh_2023_TransformerBasedLanguageModel} and \citet{michaelov_2025_LanguageModelBehavioral}, testing whether it is indeed the case that the correlation between language model surprisal and reading time peaks when the model's predictions most reflect $n$-grams.

\subsection{Method}
We use the same reading time metrics as in Experiment 1. Given our aim to explain results such as those in \citet{oh_2023_TransformerBasedLanguageModel}, we use the Pythia  models \citep{biderman_2023_PythiaSuiteAnalyzing} to calculate neural language model surprisal. These are a set of 10 autoregressive transformer language models ranging in size from 14 million parameters to 12 billion parameters, and trained on the 300 billion tokens of the aforementioned Pile text corpus, with checkpoints provided at various points over the course of training. Thus, it is possible to investigate how language models surprisal correlates with reading time both as a function of parameters and training tokens. Since the models are trained on the Pile, we use the unigram, bigram, and trigram surprisals calculated on the Pile in Experiment 1.

\subsection{Results}
First, we compare the trajectory of the correlation between Pythia surprisal and reading time to that of the correlation between Pythia surprisal and $n$-gram surprisal, which we present in \Cref{fig:corr_provo}A. The degree to which First Fixation Duration is correlated with Pythia surprisal shows a highly similar time course to trigram surprisal, while the time course for the other metrics of reading time more closely resemble those of unigram and bigram surprisal. 

To further compare the synchronicity between correlation to these variables, we look at the correlation between these correlations (\Cref{fig:corr_provo}B). As can be seen, in almost all cases, the highest correlation to reading time occurs at the point at which surprisal is most correlated with $n$-gram surprisal. The nonlinearities in the correlations (i.e., the ``V'' shapes) show that as models continue to train, their surprisal's correlation with reading time drops more slowly than the correlation with unigram surprisal (relative to the initial increase), a pattern that reduces and even reverses for higher-order $n$-gram surprisal. Thus, relationship between model surprisal's correlation to bigram surprisal and to First Pass Duration and Go-Past Duration is relatively linear, and the same is true for First Fixation Duration and Total Duration in their relation to trigram surprisal. This pattern is borne out when we calculate the correlation between these correlations (see \Cref{tab:results}). 

\begin{table*}[ht]
\begin{center}
\caption{The Pearson correlation coefficient $r$ between the correlation between language model surprisal and $n$-gram surprisal and the correlation between language model surprisal and First Fixation Duration (FFD), First Pass Duration (FPD), Go-Past Duration (GPD), and Total Duration (TD); with the highest bolded.}
\label{tab:results}
\vskip 0.12in
\begin{tabular}{@{}llllllllllll@{}}
\toprule
\textbf{} &
  \multicolumn{1}{c}{\textbf{}} &
  \multicolumn{5}{c}{\textbf{GECO}} &
  \textbf{} &
  \multicolumn{4}{c}{\textbf{Provo}} \\ \cmidrule(lr){3-7} \cmidrule(l){9-12} 
\textbf{Family} &
  \textbf{N-gram} &
  \textbf{FFD} &
  \textbf{FPD} &
  \textbf{SGPD} &
  \textbf{GPD} &
  \textbf{TD} &
  \textbf{} &
  \textbf{FFD} &
  \textbf{FPD} &
  \textbf{GPD} &
  \textbf{TD} \\ \midrule
Gemstone &
  Unigram Surprisal &
  0.964 &
  0.981 &
  \textbf{0.984} &
  \textbf{0.990} &
  0.980 &
   &
  0.884 &
  0.980 &
  0.955 &
  0.949 \\
Gemstone &
  Bigram Surprisal &
  0.942 &
  \textbf{0.986} &
  \textbf{0.984} &
  0.970 &
  0.966 &
   &
  \textbf{0.896} &
  \textbf{0.986} &
  \textbf{0.972} &
  \textbf{0.966} \\
Gemstone &
  Trigram Surprisal &
  \textbf{0.976} &
  0.970 &
  0.974 &
  0.982 &
  \textbf{0.982} &
   &
  0.886 &
  0.959 &
  0.969 &
  0.962 \\ \midrule
Open GPT-2 &
  Unigram Surprisal &
  0.564 &
  0.643 &
  0.662 &
  0.771 &
  0.646 &
   &
  0.511 &
  0.772 &
  0.723 &
  0.677 \\
Open GPT-2 &
  Bigram Surprisal &
  0.918 &
  \textbf{0.974} &
  \textbf{0.976} &
  0.938 &
  0.964 &
   &
  0.955 &
  \textbf{0.989} &
  \textbf{0.973} &
  \textbf{0.988} \\
Open GPT-2 &
  Trigram Surprisal &
  \textbf{0.95} &
  0.945 &
  0.953 &
  \textbf{0.969} &
  \textbf{0.966} &
   &
  \textbf{0.991} &
  0.918 &
  0.908 &
  0.963 \\ \midrule
Pythia &
  Unigram Surprisal &
  0.238 &
  0.377 &
  0.409 &
  0.671 &
  0.355 &
   &
  0.478 &
  0.743 &
  0.808 &
  0.622 \\
Pythia &
  Bigram Surprisal &
  0.878 &
  \textbf{0.967} &
  \textbf{0.971} &
  0.933 &
  0.945 &
   &
  0.925 &
  \textbf{0.993} &
  \textbf{0.984} &
  0.973 \\
Pythia &
  Trigram Surprisal &
  \textbf{0.931} &
  0.957 &
  0.964 &
  \textbf{0.949} &
  \textbf{0.967} &
   &
  \textbf{0.993} &
  0.952 &
  0.899 &
  \textbf{0.986} \\ \bottomrule
\end{tabular}
\end{center}
\end{table*}

\subsection{Discussion}

Overall, we see a general trend where the correlation between language model surprisal and First Pass Duration and Go-Past Duration is highly correlated with the correlation between language model surprisal and bigram surprisal. Meanwhile, for First Fixation Duration and Total Duration, we see a closer relationship with trigram surprisal. The former is surprising in light of the pattern that in the case of neural indices of processing, later responses appear to correlate with more complex relations between words and their context \citep{vanpetten_2012_PredictionLanguageComprehension,delong_2020_ComprehendingSurprisingSentences,kuperberg_2020_TaleTwoPositivities}; though it is worth noting the generally low overall correlation between all surprisal metrics and First Fixation Duration.

\section{Experiment 3}
Our results show an extremely high correlation between the extent to which language model surprisal correlates with reading time and the extent to which it correlates with $n$-gram surprisal. However, in the previous experiment, we only look at one set of language models (Pythia) and one reading time dataset (the Provo Corpus). In this experiment, we test the reliability of the results of Experiment 2 by replicating the analysis with more models on more reading time data.

\subsection{Method}
We include all the reading time data from Experiments 1--2, as well as the Ghent Eye-Tracking Corpus (GECO; \citealp{cop_2017_PresentingGECOEyetracking}), a dataset made up of 14 experimental participants reading a novel. This dataset includes an additional metric known as Selective Go-Past Duration (SGPD), a variant of go-past duration that only includes the time spent fixated on the word itself. After carrying out the same exclusions and calculating mean reading times, we carried out our analysis on data from 43,251 words.

In addition to calculating neural language model surprisal with the Pythia models (as in Experiment 2), we calculate the surprisal of all words in both reading time datasets using the Open GPT-2 \citep{karamcheti_2021_MistralJourneyReproducible} and Gemstone \citep{mcleish_2025_GemstonesModelSuitea} models. The Open GPT-2 models are trained on the OpenWebText corpus, and are an open replication of the small (124M parameter) and medium (355M parameter) GPT-2 models, with checkpoints taken over the course of training during 5 training runs (with different random seeds). Due to an error with one of these seeds \citep{hawkins_2023_ArwenCheckpointProgression}, we use the remaining 4 seeds of each model. The Gemstone models are a set of 22 models ranging from 50 million to 2 billion parameters, designed to for testing whether the width and depth of language models have an impact on performance above and beyond their contribution to total parameter count. They are trained on a subset of the Dolma corpus, with checkpoints taken over the course of training.

\subsection{Results}
The results are presented in \Cref{fig:et_both}. Across models, metrics, and datasets, we see a general trend: a language model that makes predictions that are more correlated with $n$-grams also makes predictions that are more correlated with reading time. As in Experiment 2, this is generally true to the greatest extent for bigram and trigram surprisal (see \Cref{tab:results}). Also as in Experiments 1--2, we see the lowest overall correlation between surprisal (neural language model and $n$-gram) and First Fixation Duration and Go-Past Duration---in GECO, this is more pronounced for the latter. A notable detail in \Cref{fig:et_both} is that we see mostly the same pattern across model families---the relationship between the correlation to $n$-grams and to reading time appears stable. 

\subsection{Discussion}
We observe the same general pattern as in Experiment 2, that language models that make more $n$-gram-like predictions also make predictions that are more strongly correlated with each of the reading time metrics.

\section{General Discussion}
Our results provide clear evidence that the extent to which a neural language model's predictions reflect low-order $n$-grams is correlated with the extent to which the surprisal of words calculated using it correlates with reading time. This is true across reading time metric, dataset, and language model family. Thus, the evidence is consistent with our original hypothesis---more $n$-gram-like language models show a better correlation to naturalistic reading time.

What does this tell us about human language processing? As previously noted, we do not argue for $n$-gram-based prediction as a complete account of how context shapes reading time. Nor do we argue that humans do not rely on more complex statistical relationships between words---indeed, recent work has shown that more powerful language models trained on more data other indices of language processing better than models trained on less data \citep{michaelov_2024_RevengeFallenRecurrent,michaelov_2026_BetterLanguageModels}. Nonetheless, we clearly observe that more $n$-gram-like language models make predictions that correlate better with reading time, and note that a substantial amount of previous empirical work has demonstrated a reliable link between reading time and $n$-gram surprisal, as previously discussed. Thus, taken together, our results suggest that to the extent that language statistics play a role in the reading of naturalistic text, these are best captured by $n$-grams.

The idea that first pass duration can explain the pattern in other metrics is also supported by the data analyzed in that most words are only fixated once in the first pass (Provo: 82.5\%; GECO: 88.7\%), and most words are not refixated after the first pass (Provo: 82.5\%; GECO: 83.4\%). The lower overall correlation for First Fixation Duration is likely due to the fact that in cases where there is a difference between it and First Pass Duration, it is smaller, and thus, proportionally more impacted by spillover effects from previous words. Similarly, Go-Past Duration includes the time taken to read other words in regressions from the current word.

Another question is why we see slightly higher correlations between neural language models and reading time (at their best) than between raw $n$-grams and reading time. One possibility is that neural language models simply learn a better smoothing function (recall that we use a very simple backoff scheme). Alternatively, it is possible that eye movements are sensitive to other basic properties of words that are captured by weaker neural language models but not $n$-grams, such as word-level semantic relatedness (see \citealp{michaelov_2025_LanguageModelBehavioral}). Disentangling these possibilities is a question for future work.

\clearpage
\section*{Acknowledgments}
For their discussion and helpful comments, we would like to thank the members of the Computational Psycholinguistics Laboratory at MIT, Benjamin Bergen, and the attendees of the 39th Annual Conference on Human Sentence Processing (HSP 2026). We would also like to thank Catherine Arnett for feedback on an earlier version of this manuscript, as well as the anonymous reviewers at HSP and CogSci 2026. James Michaelov was supported by a grant from the Andrew W. Mellon foundation (\#2210-13947) during the writing of this paper.

\printbibliography

\end{document}